\documentclass[runningheads,a4paper]{llncs}

\usepackage{amssymb}
\usepackage{graphicx}

\usepackage[T1]{fontenc}
\usepackage{wasysym}

\usepackage{mathptmx}       
\usepackage{helvet}         
\usepackage{courier}        
\usepackage{type1cm}        
\usepackage{url}
\urldef{\mailsa}\path|{jungyeul.park, mouna.chebbah,arnaud.martin}@univ-rennes1.fr|
\urldef{\mailsb}\path|jendoubi.siwar@etudiant.univ-rennes1.fr|
\newcommand{\keywords}[1]{\par\addvspace\baselineskip
\noindent\keywordname\enspace\ignorespaces#1}
\newcommand{\argmax}{\arg\!\max}

\begin{document}

\mainmatter  

\title{Second-order Belief Hidden Markov Models}

\titlerunning{Second-order Belief HMMs}

\author{Jungyeul Park\inst{1}, Mouna Chebbah\inst{1,2} \and Siwar Jendoubi\inst{1,2}, Arnaud Martin\inst{1}}
\authorrunning{Park et al.}
\institute{UMR 6074 IRISA, Universit\'e de Rennes1, Lannion, France\and
LARODEC Laboratory, University of Tunis, ISG Tunis, Tunisia\\
\mailsa\\
\mailsb} \maketitle

\begin{abstract}
Hidden Markov Models (HMMs) are learning methods for pattern recognition.  The probabilistic HMMs have been one of the most used techniques based on the Bayesian model. First-order probabilistic HMMs were adapted to the theory of belief functions such that Bayesian probabilities were replaced with mass functions. In this paper, we present a second-order Hidden Markov Model using belief functions. Previous works in belief HMMs have been focused on the first-order HMMs. We extend them to the second-order model.
\end{abstract}

\section{Introduction}
\label{intro}

A Hidden Markov Model (HMM) is one of the most important statistical models in machine learning \cite{rabiner:1989}. A HMM is a classifier or labeler that can assign label or class to each unit in a sequence \cite{jurafsky-martin:2008}. It has been successfully utilized over several decades in many applications for processing text and speech such as Part-of-Speech (POS) tagging \cite{kupiec:1992}, named entity recognition \cite{zhou-su:2002:ACL} and speech recognition \cite{huang-ariki-jack:1990}. However, such works in the early part of the period are mainly based on first-order HMMs. As a matter of fact, the assumption in the first-order HMM, where the state transition and output observation depend only on one previous state, does not exactly match with the real applications \cite{lmlee-jclee:2006}. Therefore, they require a number of sophistications. For example, even though the first-order HMM for POS tagging in early 1990s performs reasonably well, it captures a more limited amount of the contextual information than is available \cite{thede-harper:1999:ACL}. As consequence, most modern statistical POS taggers use a second-order model \cite{brants:2000:ANLP}.

Uncertainty theories can be integrated in statistical models such as HMMs: The probability theory has been used to classify units in a sequence with the Bayesian model. Then, the theory of belief functions is employed to this statistical model because the fusion proposed in this theory simplifies computations of \textit{a posteriori} distributions of hidden data in Markov models. 
This theory can provide rules to combine evidences from different sources to reach a certain level of belief \cite{shafer:1976,yager:1987,smets-kennes:1994,dubois-prade:1999,smets:2007}. 
Belief HMMs introduced in \cite{fouque-appriou-pieczynski:2000,lanchantin-pieczynski:2005,pieczynski:2007,ramasso:2007,soubaras:2010,ramasso-denoeux-zerhouni:2012,boudaren-et-al:2012,jendoubi-yaghlane-martin:2013,ramasso-denoeux:2014}, use combination rules proposed in the framework of the theory of belief functions. This paper is an extension of previous ideas for second-order belief HMMs. For the current work, we focus on explaining a second-order model. However, the proposed method can be easily extended to higher-order models. 

This paper is organized as follows: In Sections \ref{1st-proba} and \ref{2nd-proba}, we detail probabilistic HMMs for the problem of POS tagging where HMMs have been widely used. Then, we describe the first-order belief HMM in Section \ref{first-belief-hmm}. Finally, before concluding, we propose the second-order belief HMM.

%


\section{First-order probabilistic HMMs}
\label{1st-proba}
POS tagging is a task of finding the most probable {estimated} sequence of $n$ tags given the observation sequence of $v$ words. According to \cite{rabiner:1989}, a first-order {probabilistic} HMM can be characterized as follows:

\vspace{0.1cm}
\begin{tabular}{ ll } 
$N$ & The number of states in a model $S_t=\{s^{t}_{1}, s^{t}_{2}, \cdots  s^{t}_{N}\}$ at the time $t$.\\
$M$ & The number of distinct observation symbols.
$V=\{v_{1}, v_{2}, \cdots , v_{M}\}$.\\
$A=\{a_{ij}\}$ &  The set of $N$ transition probability distributions. \\
$B=\{b_{j}(o_{t})\}$ & The observation probability distributions in state $j$.\\
$\pi=\{\pi_{i}\}$ & The initial probability distribution.\\ 
\end{tabular} 
\vspace{0.1cm}

\begin{figure} [t]
\begin{center}
\includegraphics[scale=0.78]{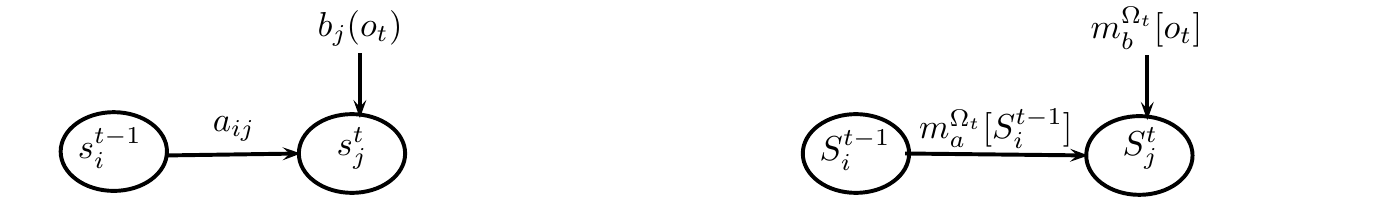} 
\end{center}
\caption{First-order probabilistic and belief HMMs}
\label{1st-hmm}
\end{figure}

Figure \ref{1st-hmm} illustrates the first-order probabilistic HMM allowing to estimate the probability of the sequence $s^{t-1}_{i}$ and $s^{t}_{j}$ where $a_{ij}$ is the transition probability from $s^{t-1}_{i}$ to $s^{t}_{j}$ and $b_{j}(o_{t})$ is the observation probability on the state $s^{t}_{j}$. Regarding POS tagging, the number of possible POS tags that are hidden states $S_{t}$ of the HMM is $N$. The number of words in the lexicons $V$ is $M$. The transition probability $a_{ij}$ is the probability that the model moves from one tag $s^{t-1}_{i}$ to another tag $s^{t}_{j}$. This probability can be estimated using a training data set in supervised learning for the HMM. The probability of a current POS tag appearing in the first-order HMM depends only on the previous tag. In general, first-order probabilistic HMMs should be characterized by three fundamental problems as follows \cite{rabiner:1989}: 

\begin{itemize}
\item {Likelihood}: Given a set of transition probability distributions $A$, an observation sequence $O=o_{1}, o_{2}, \cdots, o_{T}$ and its observation probability distribution $B$, how do we determine the likelihood $P(O|A,B)$? The first-order model relies on only one observation where $b_{j}(o_{t})=P(o_{j}|s^{t}_{j})$ and the transition probability based on one previous tag where $a_{ij}$ = $P(s^{t}_{j}|s^{t-1}_{i})$. Using the forward path probability, the likelihood $\alpha_{t}(j)$ of a given state $s^{t}_{j}$ can be computed by using the likelihood $\alpha_{t-1}(i)$ of the previous state $s^{t-1}_{i}$ as described below:

\begin{equation}
\alpha_{t}(j)=\sum_{i}{\alpha_{t-1}(i)a_{ij}b_{j}(o_{t})}
\end{equation}

\item {Decoding}: Given a set of transition probability distributions $A$, an observation sequence $O=o_{1}, o_{2}, \cdots, o_{T}$ and its observation probability distribution $B$, how do we discover the best hidden state sequence? The Viterbi algorithm is widely used for calculating the most likely tag sequence for the decoding problem. The Viterbi algorithm can calculate the most probable path $\delta_{t}(j)$ which contains the sequence of $\psi_{t}(j)$. It can select the path that maximizes the likelihood of the sequence as described below:

\begin{equation}
\begin{tabular}{ l}
$\delta_{t}(j)=\max{\delta_{t-1}(i)a_{ij}b_{j}(o_{t})}$\\
$\psi_{t}(j)=\argmax{\psi_{t-1}(i)a_{ij}}$\\
\end{tabular} 
\end{equation}

\item {Learning}: Given an observation sequence $O=o_{1}, o_{2}, \cdots, o_{T}$ and a set of states $S=\{s^{t}_{1}, s^{t}_{2}, \cdots , s^{t}_{N}\}$, how do we learn the HMM parameters for $A$ and $B$?
The parameter learning task usually uses the Baum-Welch algorithm which is a special case of the Expectation-Maximization (EM) algorithm.
\end{itemize}

In this paper, we focus on the likelihood and decoding problems by assuming a supervised learning paradigm where labeled training data are already available. 

\section{Second-order probabilistic HMMs}
\label{2nd-proba}
\begin{figure} [t]
\begin{center}
\includegraphics[scale=0.78]{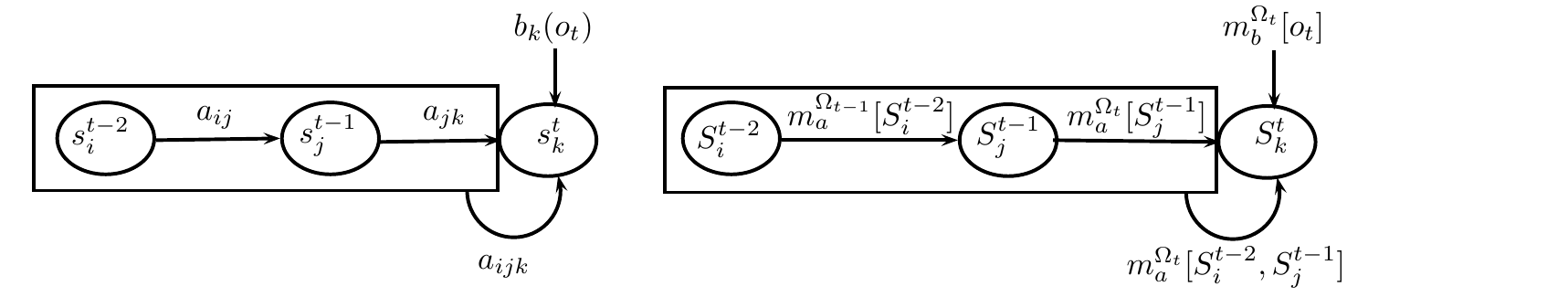} 
\end{center} 
\caption{Second-order probabilistic and belief HMMs}
\label{2nd-hmm}
\end{figure}

Now, we explain the extension of the first-order model to a \textit{trigram}\footnote{The trigram is the sequence of three elements, \textit{i.e.} three states in our case.} in the second-order model. Figure \ref{2nd-hmm} illustrates the second-order probabilistic HMM allowing to estimate the probability of the sequence of three states $s^{t-2}_{i}$, $s^{t-1}_{j}$ and $s^{t}_{k}$ where $a_{ijk}$ is the transition probability from $s^{t-2}_{i}$ and $s^{t-1}_{j}$ to $s^{t}_{k}$, and $b_{k}(o_{t})$ is the observation probability on the state $s^{t}_{k}$. Therefore, the second-order probabilistic HMM is characterized by three fundamental problems as follows: 

\begin{itemize}
\item {Likelihood}:  The second-order model relies on one observation $b_{k}(o_{t})$. Unlike the first-order model, the transition probability is based on two previous tags where $a_{ijk}$ = $P(s^{t}_{k}|s^{t-2}_{i}$, $s^{t-1}_{j})$ as described below:

\begin{equation}
\alpha_{t}(k)=\sum_{j}{\alpha_{t-1}(j)a_{ijk}b_{k}(o_{t})}
\end{equation}

However, it will be more difficult to find a sequence of three tags than a sequence of two tags. Any particular sequence of tags $s^{t-2}_{i}$, $s^{t-1}_{j}$, $s^{t}_{k}$ that occurs in the test set may simply never have occurred in the training set  because of data sparsity \cite{jurafsky-martin:2008}. Therefore, a method for estimating $P(s^{t}_{k}|s^{t-2}_{i}, s^{t-1}_{j})$, even if the sequence $s^{t-2}_{i}$, $s^{t-1}_{j}$, $s^{t}_{k}$ never occurs, is required. The simplest method to solve this problem is to combine the trigram $\hat{P}(s^{t}_{k}|s^{t-2}_{i}, s^{t-1}_{j})$, the bigram $\hat{P}(s^{t}_{k}|s^{t-1}_{j})$, and even the unigram $\hat{P}(s^{t}_{k})$ probabilities \cite{brants:2000:ANLP}:

\begin{equation}
P(s^{t}_{k}|s^{t-2}_{i}, s^{t-1}_{j}) = \lambda_{1}\hat{P}(s^{t}_{k}|s^{t-2}_{i}, s^{t-1}_{j}) + \lambda_{2}\hat{P}(s^{t}_{k}|s^{t-1}_{j}) + \lambda_{3}\hat{P}(s^{t}_{k})
\end{equation}

Note that $\hat{P}$ is the maximum likelihood probabilities which are derived from the relative frequencies of the sequence of tags. 
Values of $\lambda$ are such that $\lambda_{1}+\lambda_{2}+\lambda_{3}=1$ and they can be estimated by the \textit{deleted interpolation} algorithm \cite{brants:2000:ANLP}. Otherwise, \cite{thede-harper:1999:ACL} describes a different method for values of $\lambda$ as below:

\begin{equation}
\begin{tabular}{ ll }
$\lambda_{1} = $ & $k_{3}$ \\
$\lambda_{2} = $ & $(1-k_{3}) \cdot k_{2}$ \\
$\lambda_{3} = $ & $(1-k_{3}) \cdot (1-k_{2})$ \\
\end{tabular} 
\end{equation}

\noindent where {$k_{2} = \frac{log(C(s^{t-1}_{j}, s^{t}_{k})+1)+1}{log(C(s^{t-1}_{j}, s^{t}_{k})+1)+2}$}, {$k_{3} = \frac{log(C(s^{t-2}_{i}, s^{t-1}_{j}, s^{t}_{k})+1)+1}{log(C(s^{t-2}_{i}, s^{t-1}_{j}, s^{t}_{k})+1)+2}$}, and $C(s^{t-2}_{i}, s^{t-1}_{j}, s^{t}_{k})$ is the frequency of a sequence $s^{t-2}_{i}, s^{t-1}_{j}, s^{t}_{k}$ in the training data. Note that $\lambda_{1}+\lambda_{2}+\lambda_{3}$ is not always equal to one in \cite{thede-harper:1999:ACL}. The likelihood of the observation probability for the second-order model uses $B$ where $b_{k}(o_{t})=P(o_{k}|s^{t}_{k}, s^{t-1}_{j})$. 

\item {Decoding}: For second-order model we require a different Viterbi algorithm. For a given state $s$ at the time $t$, it would be redefined as follows \cite{thede-harper:1999:ACL}:

\begin{equation}
\begin{tabular}{ l }
$\delta_{t}(k)=\max{\delta_{t-1}(j)a_{ijk}b_{k}(o_{t})}$ \\
\ \ \ \ \ where $\delta_{t}(j)=\max{P(s^{1}, s^{2}, \cdots , s^{t-1}=s_{i}, s^{t}=s_{j},o_{1}, o_{2}, \cdots , o_{t})}$\\
$\psi_{t}(k)=\argmax{\psi_{t-1}(j)a_{ijk}}$\\
\ \ \ \ \ where $\psi_{t}(k)=\argmax{P(s^{1}, s^{2}, \cdots , s^{t-1}=s_{i}, s^{t}=s_{j},o_{1}, o_{2}, \cdots , o_{t})}$\\
\end{tabular} 
\end{equation}

\item {Learning}: The problem of learning would be similar to the first-order model except that parameters $A$ and $B$ are different.
\end{itemize} 

With respect to performance measures, different transition probability distributions in \cite{brants:2000:ANLP} and \cite{thede-harper:1999:ACL} obtain 97.0\% and 97.09\% tagging accuracy for known words, respectively for the same data (the Penn Treebank corpus). Even though probabilistic HMMs perform reasonably well, belief HMMs  can learn better under certain conditions on observations \cite{jendoubi-yaghlane-martin:2013}. 

\section{First-order Belief HMMs}
\label{first-belief-hmm}

In probabilistic HMMs, $A$ and $B$ are probabilities estimated from the training data. However, $A$ and $B$ in belief HMMs are mass functions (bbas) 
\cite{ramasso:2007,jendoubi-yaghlane-martin:2013}.
According to previous works on belief HMMs, a first-order HMM using belief functions can be characterized as follows\footnote{In the model $\Omega_{t}$, $S^{t}$ are focal elements}:

\vspace{0.1cm}
\begin{tabular}{ ll }
$N$ & The number of states in a model $\Omega_{t}=\{S^{t}_{1}, S^{t}_{2}, \cdots , S^{t}_{N}\}$.\\
$M$ & The number of distinct observation symbols $V$. \\
$A=\{m^{\Omega_{t}}_{a}[S^{t-1}_{i}](S^{t}_{j})\}$ &  The set of conditional bbas to all possible subsets of states. \\
$B=\{m^{\Omega_{t}}_{b}[o_{t}](S^{t}_{j})\}$ & The set of bbas according to all possible observations $O_{t}$. \\
$\pi=\{m^{\Omega_{1}}_{\pi}(S^{\Omega_{1}}_{i})\}$ & The bba defined for the the initial state.\\
\end{tabular} 
\vspace{0.1cm}

Difference between the first-order probabilistic and belief HMMs is presented in Figure \ref{1st-hmm}, the transition and observation probabilities in belief HMMs are described as mass functions. Therefore, we can replace $a_{ij}$ by $m^{\Omega_{t}}_{a}[S^{t-1}_{i}](S^{t}_{j})$ and $b_{j}(o_{t})$ by $m^{\Omega_{t}}_{b}[o_{t}](S^{t}_{j})$. The set $\Omega_{t}$ has been used to denote states for HMMs using belief functions \cite{ramasso:2007,jendoubi-yaghlane-martin:2013}. Note that $s^{t}_{i}$ is the single state for probabilistic HMMs and $S^{t}_{i}$ is the multi-valued state for belief HMMs. First-order belief HMMs should also be characterized by three fundamental problems as follows:

\begin{itemize}
\item {Likelihood}: The likelihood problem in belief HMMs is not solved by \textit{likelihood}, but by using the combination. The first-order belief model relies on (\textit{i}) only one observation $m^{\Omega_{t}}_{b}[o_{t}](S^{t}_{j})$ and (\textit{ii}) a transition conditional mass function based on one previous tag $m^{\Omega_{t}}_{a}[S^{t-1}_{i}](S^{t}_{j})$. Mass functions of sets $A$ and $B$ are combined using the Disjunctive Rule of Combination (DRC) for the forward propagation and the Generalized Bayesian Theorem (GBT) for the backward propagation \cite{smets:1993}. Using the forward path propagation, the mass function of a given state $S^{t}_{j}$ can be computed as the combination of mass functions on the observation and the transition as described below:

\begin{equation}
q^{\Omega_{t}}_{\alpha}(S^{t}_{j})=\sum{m^{\Omega_{t-1}}_{\alpha}(S^{t-1}_{i}) \cdot q^{\Omega_{t}}_{a}[S^{t-1}_{i}](S^{t}_{j}) \cdot q^{\Omega_{t}}_{b}(S^{t}_{j})}
\end{equation}

Note that the mass function of the given state $S^{t}_{j}$ is derived from the commonality function $q^{\Omega_{t}}_{\alpha}$. 

\item {Decoding}: Several solutions have been proposed to extend the Viterbi algorithm to the theory of belief functions \cite{ramasso:2007,serir-ramasso-zerhouni:2011,ramasso:2011}. Such solutions maximize the plausibility of the state sequence. In fact, the \textit{credal} Viterbi algorithm starts from the first observation and estimates the commonality distribution of each observation until reaching the last state. For each state $S^{t}_{j}$, the estimated commonality distribution ($q^{\Omega_{t}}_{\delta}(S^{t}_{j})$) is converted back to a mass function that is conditioned on the previous state. Then, we apply the \textit{pignistic} transform to make a decision about the current state ($\psi_{t}(s^{t}_{j})$):

\begin{equation}
\begin{tabular}{ l }
$q^{\Omega_{t}}_{\delta}(S^{t}_{j})=\sum_{S^{t-1}_{i}\subseteq A^{t-1}   }{m^{\Omega_{t-1}}_{\delta}(S^{t-1}_{i}) \cdot q^{\Omega_{t}}_{a}[S^{t-1}_{i}](S^{t}_{j}) \cdot q^{\Omega_{t}}_{b}(S^{t}_{j})}$ \\
$\psi_{t}(s^{t}_{j})=\argmax_{S^{t-1}_{i} \in \Omega_{t-1} }  {(1-m^{\Omega_{t}}_{\delta}[S^{t-1}_{i}](\emptyset)) \cdot  P_{t} [S^{t-1}_{i}] (S^{t}_{j})}$\\
\end{tabular}
\end{equation}

\noindent where $A^{t}=\cup_{S^{t-1}_{j}\in \Omega_{t}}\psi_{t}(S^{t}_{j})$ \cite{ramasso:2007}.
\item {Learning}: Instead of the traditional EM algorithm, we can use the $E^{2}M$ algorithm for the belief HMM \cite{ramasso-denoeux:2014}.

\end{itemize}

To build belief functions from what we learned using probabilities in the previous section, we can employ the least commitment principle by using the inverse pignistic transform \cite{sudano:2002,aregui-denoeux:2008}. 

\section{Second-order Belief HMMs}
Like the first-order belief HMM, $N$, $M$, $B$ and $\pi$ are similarly defined in the second-order HMM. The set $A$ is quite different and is defined as follows:

\begin{equation}
A=\{m^{\Omega_{t}}_{a}[S^{t-2}_{i}, S^{t-1}_{j}](S^{t}_{k})\}
\end{equation}

\noindent where $A$ is the set of conditional bbas to all possible subsets of states based on the two previous states. Second-order belief HMMs should also be characterized by three fundamental problems as follows: 

\begin{itemize}
\item {Likelihood}: The second-order belief model relies on one observation $m^{\Omega_{t}}_{b}[o_{t}](S^{t}_{k})$ in a state $S_{k}$ at time $t$ and the transition conditional mass function based on two previous states $S^{t-2}_{i}$ and $S^{t-1}_{j}$, defined by $m^{\Omega_{t}}_{a}[S^{t-2}_{i}, S^{t-1}_{j}](S^{t}_{k})$. Using the forward path propagation, the mass function of a given state $S^{t}_{k}$ can be computed as the disjunctive combination (DRC) of mass functions on the transition $m^{\Omega_{t}}_{a}[S^{t-2}_{i}, S^{t-1}_{j}](S^{t}_{k})$ and the observation $m^{\Omega_{t}}_{b}(S^{t}_{k})$ as described below:

\begin{equation}
q^{\Omega_{t}}_{\alpha}(S^{t}_{k})=\sum{m^{\Omega_{t-1}}_{\alpha}(S^{t-1}_{j}) \cdot q^{\Omega_{t}}_{a}[S^{t-2}_{i}, S^{t-1}_{j}](S^{t}_{k}) \cdot q^{\Omega_{t}}_{b}(S^{t}_{k})}
\end{equation}

\noindent where $q^{\Omega_{t}}_{a}[S^{t-2}_{i}, S^{t-1}_{j}](S^{t}_{k})$ is the commonality function derived from the conjunctive combination of mass functions of two previous transitions. The conjunctive combination is used to have the conjunction of observations on previous two states $S^{t-2}_{i}$ and $S^{t-1}_{j}$. 

 The combined mass function $m^{\Omega_{t}}_{a}[S^{t-2}_{i}, S^{t-1}_{j}](S^{t}_{k})$ of two transitions $m^{\Omega_{t-1}}_{a}[S^{t-2}_{i}](S^{t-1}_{j})$ and $m^{\Omega_{t}}_{a}[S^{t-1}_{j}](S^{t}_{k})$ is defined as follows: 

\begin{equation}
\begin{tabular}{ l }
$m^{\Omega_{t}}_{a}[S^{t-2}_{i}, S^{t-1}_{j}](S^{t}_{k})=m^{\Omega_{t-1}}_{a}[S^{t-2}_{i}](S^{t-1}_{j})$ {{\textcircled{{\scriptsize $\cup$}}}} $m^{\Omega_{t}}_{a}[S^{t-1}_{j}](S^{t}_{k})$\\
\end{tabular}
\end{equation}

The conjunctive combination is required to obtain the conjunction of both transitions. 
Note that the mass function of the given state $S^{t}_{k}$ is derived from the commonality function $q^{\Omega_{t}}_{\alpha}$. We use DRC with commonality functions like in \cite{ramasso:2007}. 
Note that the observation only on one previous state is taken into account in the first-order belief HMM, but the conjunction of observations on two previous states is considered in the second-order belief HMM.

\item {Decoding}: We accept our assumption of the first-order belief HMM for the second-order model. Similarly to the first-order belief HMM, we propose a solution that maximizes the plausibility of the state sequence. The credal Viterbi algorithm estimates the commonality distribution of each observation from the first observation till the final state. For each state $S^{t}_{k}$, the estimated commonality distribution ($q^{\Omega_{t}}_{\delta}(S^{t}_{k})$) is converted back to a mass function that is conditioned on a mass function of the two previous states. This mass function is the conjunctive combination of mass functions of the two previous states. Then, we apply the pignistic transform to make a decision about the current state ($\psi_{t}(s^{t}_{j})$) as before:

\begin{equation}
\begin{tabular}{ l }
$q^{\Omega_{t}}_{\delta}(S^{t}_{k})=\sum_{S^{t-1}_{j}\subseteq A^{t-1}   }{m^{\Omega_{t-1}}_{\delta}(S^{t-1}_{j}) \cdot q^{\Omega_{t}}_{a}[S^{t-2}_{i},S^{t-1}_{j}](S^{t}_{k}) \cdot q^{\Omega_{t}}_{b}(S^{t}_{k})}$ \\
$\psi_{t}(s^{t}_{k})=\argmax_{S^{t-1}_{j} \in \Omega_{t-1} }  {(1-m^{\Omega_{t}}_{\delta}[S^{t-1}_{j}](\emptyset)) \cdot  P_{t} [S^{t-2}_{i},S^{t-1}_{j}] (S^{t}_{k})}$\\
\end{tabular}
\end{equation}

\item {Learning}: Like the first-order belief model, we can still use the $E^{2}M$ algorithm for the belief HMM \cite{ramasso-denoeux:2014}.

\end{itemize}

Since the combination of mass functions in the belief HMM is required where the previous observation is already considered in the set of conditional bbas $m^{\Omega_{t}}_{a}[S^{t-2}_{i}, S^{t-1}_{j}]$, we do not need to refine the observation probability for the second-order model as in the second-order probabilistic model.

\section{Conclusion and future perspectives}
\label{conclusion}
The problem of POS tagging has been considered as one of the most important tasks for natural language processing systems. We described such a problem based on HMMs and tried to apply our idea to the theory of belief functions. We extended previous works on belief HMMs to the second-order model. Using the proposed method, we will be able to easily extend the higher-order model for belief HMMs. Some technical aspects still remain to be considered. Robust implementation for belief HMMs are required where in general we can find over one million observations in the training data to deal with the problem of POS tagging. As described before, the choice of inverse pignistic transforms would be empirically verified.\footnote{For example, \cite{fayad-cherfaoui:2009} used the inverse pignistic transform in \cite{sudano:2002} to calculate belief functions from Bayesian probability functions. As matter of fact, the problem of POS tagging can be normalized and inverse pignistic transforms in \cite{sudano:2002} did not propose the case for $m(\emptyset)$. } We are planning to implement these technical aspects in near future.

The current work is described to rely on a supervised learning paradigm from labeled training data. Actually, the forward-backward algorithm in HMMs can do completely unsupervised learning. However, it is well known that EM performs poorly in unsupervised induction of linguistic structure because it tends to assign relatively equal numbers of tokens to each hidden state \cite{johnson:2007:EMNLP-CoNLL2007}.\footnote{The actual distribution of POS tags would be {highly skewed} as in heavy-tail distributions.} Therefore, the initial conditions can be very important.  Since the theory of belief functions can take into consideration of uncertainty and imprecision, especially for the lack of data, we might obtain a better model using belief functions on an unsupervised learning paradigm.

\bibliographystyle{splncs}
\bibliography{belief2014.bib}

\end{document}